\documentclass[fleqn,10pt]{wlscirep}
\usepackage[utf8]{inputenc}
\usepackage[T1]{fontenc}
\usepackage{bm,comment}
\usepackage{xcolor}
\usepackage{graphicx}
\usepackage{amsmath}
\usepackage{tikz}
\usetikzlibrary{shapes.geometric, arrows.meta, calc, fit, positioning, decorations.markings, decorations.pathmorphing, backgrounds} 

\newcommand{\Heven}{H_{\mathrm{even}}}
\newcommand{\Hodd}{H_{\mathrm{odd}}}

\title{The Shape of Sight: A Homological Framework for Unifying Visual Perception}

\author[1,*]{Xin Li}
\affil[1]{Department of Computer Science, University at Albany, Albany, NY 12222}

\affil[*]{e-mail: xli48@albany.edu}

\begin{abstract}
Visual perception, the brain's construction of a stable world from sensory data, faces several long-standing, fundamental challenges. These include the inverse problem (inferring 3D structure from ambiguous 2D inputs), the binding problem (integrating disparate features into coherent objects), and object invariance (recognizing items across transformations). While often studied separately, these problems have resisted a single, unifying computational framework. In this perspective, we propose a homological framework for visual perception. We argue that the brain's latent representations are governed by their topological parity. This parity interpretation functionally separates homological structures into two distinct classes: 1) Even-dimensional homology ($\Heven$) acts as static, integrative scaffolds. These structures bind context and content into ``wholes'' or ``what'', serving as the stable, resonant cavities for perceptual objects; 2) Odd-dimensional homology ($\Hodd$) acts as dynamic, recurrent flows. These structures represent paths, transformations, and self-sustaining ``traces'' or ``where'' that navigate the perceptual landscape. This scaffold-and-flow model is supported by the ventral-dorsal pathway separation and provides a unified solution to three core problems in visual perception. The binding problem is solved by the even-dimensional scaffolds, which act as resonant cavities that integrate features into a stable, coherent percept. Object invariance, as hypothesized by manifold unfolding, is the mathematical separation of a scaffold (the object's identity, $\Heven$) from the flows (identity-preserving transformations, $\Hodd$). Finally, the inverse problem is solved as a best-fit resonance: the ambiguous 2D input (a 2-chain) excites the most compatible 3D scaffold ($b_2$) from the brain's pre-built library of latent wholes. Homological parity hypothesis recasts visual perception not as a linear computation, but as a dynamic interaction between stable, integrative structures and the recurrent, self-sustaining flows that run on them. This perspective offers a new mathematical foundation for linking neural dynamics to perception and cognition.
\end{abstract}
\begin{document}

\flushbottom
\maketitle

\thispagestyle{empty}

\section*{Introduction}

The human visual system achieves a remarkable feat: the near-instantaneous \emph{construction} of a stable, three-dimensional world from a pair of flat, ambiguous retinal projections \cite{marr2010vision}. This process feels effortless, yet it belies a cascade of computational challenges that remain among the most profound open questions in neuroscience \cite{dicarlo2012does}. How does the brain solve an ill-posed inverse problem, bind disparate features into seamless objects, and recognize those objects from any angle, all at once?
For decades, these challenges have largely been studied in isolation, fragmenting our understanding of perception. They include the inverse Problem, the ill-posed task of inferring a unique 3D source from an infinite set of 2D possibilities \cite{von1867handbuch}; the binding problem, the question of how disparate features like color, form, and motion, processed in parallel neural pathways, are unified into a single, coherent percept \cite{treisman1996binding}; and object invariance, the capacity to recognize an object's identity despite drastic changes in viewpoint, illumination, or scale \cite{logothetis1995psychophysical}. 

In this perspective, we propose a new conceptual framework for visual perception based on algebraic topology, specifically Euler's characteristics in homological theory \cite{hatcher2005algebraic}. We posit a \textbf{homological parity hypothesis}, which asserts that the brain's latent representations are functionally and structurally segregated by their topological dimension. This parity segregates computation into two distinct classes (Fig. 1):
1) \textbf{even-dimensional homology ($\Heven$)} forms static, integrative \emph{scaffolds}. These structures act as the resonant cavities or templates that bind content and context into coherent, stable wholes;
2) \textbf{odd-dimensional homology ($\Hodd$)} forms dynamic, recurrent \emph{flows}. These structures represent recurrent traces or paths of transformation for self-reference and temporal evolution.

The proposed scaffold-and-flow architecture provides a common mathematical language for solving three challenging problems in vision. The binding problem \cite{treisman1996binding} is resolved by parity sorting as an act of resonance: features are bound when they jointly excite a stable, even-dimensional scaffold. The invariance of object recognition \cite{dicarlo2007untangling} is understood as parity \emph{separation} of these components: recognition is the act of isolating a stable scaffold (the object's identity, $\Heven$) from the flow that acts upon it (the transformation, $\Hodd$). Finally, the inverse problem \cite{blake1987visual} is solved by a low-cost verification mechanism without combinatorial explosion, where the 2D input selects the best-fit 3D scaffold from the brain's learned library of topological wholes. 


\begin{figure}[h!]
    \centering
\begin{tikzpicture}[
    >=Latex,
    font=\small,
    backbone/.style={line width=0.9pt},
    loop/.style={line width=0.8pt,dashed},
    flow/.style={-Latex, line width=0.8pt},
    anchorpt/.style={circle, fill=black, inner sep=0pt, minimum size=2.2pt},
    tag/.style={fill=black!4, draw, rounded corners, inner sep=2pt},
    lab/.style={font=\scriptsize, align=center}
]

\def\xL{0.5}
\def\xR{12.5}
\coordinate (S0) at (\xL,0);
\coordinate (S4) at (\xR,0);

\coordinate (S1) at ($(S0)!0.30!(S4)$);
\coordinate (S2) at ($(S0)!0.55!(S4)$);
\coordinate (S3) at ($(S0)!0.80!(S4)$);

\draw[backbone] (S0)--(S4);
\node[lab,above] at ($(S0)!0.5!(S4)+(3.5,-0.5)$)
  {$\sigma$ (even-dimensional scaffold: shared, reusable backbone)};

\node[lab,anchor=south west] at ($(S0)+(-0.15,0.20)$) {(a) scaffolds};

\node[anchorpt] (A1) at (S1) {};
\node[anchorpt] (A2) at (S2) {};
\node[anchorpt] (A3) at (S3) {};

\path let \p1=(A1) in coordinate (B1L) at (\x1-0.9,1.05) coordinate (B1R) at (\x1+0.9,1.05);
\draw[loop] (A1) .. controls ($(A1)+(0,0.55)$) and (B1L) .. ($(A1)+(0,1.05)$)
            .. controls (B1R) and ($(A1)+(0,0.55)$) .. (A1);
\node[lab,above] at ($(A1)+(0,1.25)$) {$b_{1}$};

\path let \p2=(A2) in coordinate (B2L) at (\x2-1.15,1.35) coordinate (B2R) at (\x2+1.15,1.35);
\draw[loop] (A2) .. controls ($(A2)+(0,0.70)$) and (B2L) .. ($(A2)+(0,1.35)$)
            .. controls (B2R) and ($(A2)+(0,0.70)$) .. (A2);
\node[lab,above] at ($(A2)+(0,1.60)$) {$b_{2}$};

\path let \p3=(A3) in coordinate (B3L) at (\x3-1.00,1.10) coordinate (B3R) at (\x3+1.00,1.10);
\draw[loop] (A3) .. controls ($(A3)+(0,0.60)$) and (B3L) .. ($(A3)+(0,1.10)$)
            .. controls (B3R) and ($(A3)+(0,0.60)$) .. (A3);
\node[lab,above] at ($(A3)+(0,1.30)$) {$b_{3}$};

\node[lab,anchor=south] at ($(A2)+(0,2.0)$) {(b) flows};

\draw[flow] ($(S0)!0.42!(S2)$) -- node[lab,below,xshift=-50pt,yshift=-1pt]
  {$\mathcal{R}$: re-enter scaffold $\sigma$} ($(S0)!0.58!(S2)$);

\draw[flow,->] ($(A2)+(-0.25,-0.08)$)
  to[bend left=18] ($(A2)+(0.05,0.80)$);
\node[lab,left] at ($(A2)+(-0.35,0.45)$) {$\mathcal{F}$};

\draw[flow,->] ($(A2)+(0.05,0.80)$)
  to[bend left=40] ($(A2)+(0.25,0.00)$);
\node[lab,right] at ($(A2)+(0.35,0.45)$) {$\mathcal{R}$};

\draw[flow] ($(A1)+(0.00,0.95)$) arc[start angle=90,end angle=-230,radius=0.95cm];

\node[tag, below=10mm of $(S0)!0.5!(S4)$] (leg) {%
\begin{minipage}{0.94\linewidth}
\footnotesize
\textbf{Legend.} \textbf{(a) Even-dimensional homology as scaffolds.}
$\sigma$ denotes a shared, low-entropy backbone (e.g. classes in $H_{0},H_{2}$)
that binds and integrates content across contexts. 
\textbf{(b) Odd-dimensional homology as flows.}
$b_k$ denote admissible recurrent loops (e.g. classes in $H_{1},H_{3}$) that ride
on $\sigma$ and encode invariants over time.
Homological parity separates \emph{structure} (even, scaffolds) from
\emph{dynamics} (odd, flows), while $\partial^{2}=0$ enforces re-closure of
flows onto their supporting scaffolds.
\end{minipage}
};

\node[lab,anchor=west] at ($(S4)+(-1.2,0.95)$)
  {even: scaffolds,\; odd: flows,\; $\partial^{2}=0$};

\end{tikzpicture}

    \caption{\textbf{The Structural Visualization of Homological Parity Hypothesis.} (a) Even-dimensional homology provides stable, integrative structures (scaffolds) that solve binding and inference. (b) Odd-dimensional homology provides dynamic, recurrent paths (flows) that solve invariance and temporal persistence.}
    \label{fig:fig1}
\end{figure}

\section*{The Homological Parity Hypothesis}

To build a new foundation for visual perception, we must first establish its mathematical ground rules. We propose that the brain's latent computational structure, like all physical and informational systems, adheres to the fundamental constraints of topology \cite{nakahara2018geometry}. These constraints provide a deep functional logic for why the brain would segregate its computations by parity.

\subsection*{From Consistency ($\partial^2=0$) to Conservation ($\chi$)}

At the heart of topology lies a single, profound rule of self-consistency: $\partial^2=0$. This is the formal statement that ``the boundary of a boundary is zero'' \cite{wheeler1990information}. Intuitively, this guarantees that all structures are closed and accounted for. For example, the 1-dimensional boundary ($\partial$) of a 2D face is a set of edges (a 1-chain). The 0-dimensional boundary ($\partial$) of this set of edges is zero, because the edges form a closed loop with no endpoints (Fig. \ref{fig:fig2}).
This rule of consistency is not just a mathematical curiosity; it is the First Law that allows a stable accounting system for structure to exist. Because the boundary of a boundary is zero ($\partial^2=0$), the boundary operator guarantees that all boundaries ($B_k$) are necessarily cycles ($Z_k$) ($B_k \subseteq Z_k$). We precisely define cycles ($Z_k$) as data with no boundary, and boundaries ($B_k$) as data that is the boundary of some higher-dimensional object. The persistent, interesting structures, the holes or latent topological features, are the cycles that are not boundaries. The set of these features, found by quotienting the cycles by the boundaries, is called the homology group ($H_k = Z_k / B_k$).

The existence of this accounting system leads directly to a deep conservation law: the fundamental \textbf{Euler-Poincaré Theorem} (Theorem 2.44 \cite{hatcher2005algebraic}). This theorem states that for any given shape, the \emph{alternating sum} of its local cells (the combinatorial count $c_k$) must equal the \emph{alternating sum} of its global holes (the Betti numbers $b_k$):
$\chi = \sum_k (-1)^k c_k = \sum_k (-1)^k b_k$
As topology provides the language of the shape, the Euler characteristic ($\chi$) is a net balance or conserved quantity. Importantly, in nature, this conservation law is \emph{innately structured by parity}: it explicitly separates the world into features that add to the sum (even dimensions: $b_0, b_2, \dots$) and features that subtract from it (odd dimensions: $b_1, b_3, \dots$). This is the first mathematical hint that even- and odd-dimensional structures must have fundamentally different, opposing roles (e.g., objective vs. subjective).


\begin{figure}[h]
    \centering
    \begin{tikzpicture}[
    node distance=2cm, 
    block/.style={rectangle, draw, fill=blue!10, text width=4.5cm, minimum height=1cm, text centered, font=\sffamily\bfseries},
    concept/.style={ellipse, draw, fill=red!10, text width=3.5cm, minimum height=1cm, text centered, font=\sffamily},
    arrow/.style={-Latex, thick, draw=black!70},
    smallarrow/.style={-Latex, thin, draw=gray!70},
    mainlabel/.style={font=\sffamily\bfseries\Large, text=black!80},
    sublabel/.style={font=\sffamily\bfseries, text=black!70},
    detail/.style={font=\sffamily\small, text=gray!70},
    line/.style={draw, thick, -Latex},
    paneltitle/.style={font=\sffamily\bfseries, text=black!80},
    mathbox/.style={rectangle, draw=gray!50, dashed, fill=gray!5, inner sep=5pt},
    betti_box/.style={rectangle, fill=blue!5, rounded corners, inner sep=4pt, draw=blue!30, text width=3.5cm, text centered},
    parity_box_even/.style={rectangle, fill=green!10, rounded corners, inner sep=5pt, draw=green!30, text width=4.5cm, text centered},
    parity_box_odd/.style={rectangle, fill=purple!10, rounded corners, inner sep=5pt, draw=purple!30, text width=4.5cm, text centered},
]

\node[paneltitle, anchor=west] (panelA_title) at (-7, 5.5) {(a) The $\partial^2=0$ Rule (Consistency)};

\coordinate (A) at (-5.5, 4.5); 
\coordinate (B) at (-3.5, 4.5);
\coordinate (C) at (-3.5, 2.5);
\coordinate (D) at (-5.5, 2.5);
\draw[fill=blue!10, draw=blue!50, line width=0.8pt] (A) -- (B) -- (C) -- (D) -- cycle;
\node[text=blue!80, font=\sffamily\bfseries\small] at ($(A)!0.5!(C)$) {$c_2$ (Face)};

\draw[line width=1.5pt, -{Latex[length=2.5mm]}, red!60] (A) -- (B);
\draw[line width=1.5pt, -{Latex[length=2.5mm]}, red!60] (B) -- (C);
\draw[line width=1.5pt, -{Latex[length=2.5mm]}, red!60] (C) -- (D);
\draw[line width=1.5pt, -{Latex[length=2.5mm]}, red!60] (D) -- (A);
\node[text=red!80, font=\sffamily\bfseries\small, above right=0.1cm and 0.1cm of B] {$\partial c_2$};

\node[mathbox, text width=3.5cm, align=left] (math_exp) at (-2, 2.5) {
    $\partial c_2 = \text{Edges}$ \\
    $\partial (\partial c_2) = \text{Endpoints} = \mathbf{0}$
};
\draw[arrow, shorten >=2mm] ($(A)!0.5!(D)$) -- (math_exp.west); 

\node[betti_box] (homology_def) at (-6.5, 0) 
{ 
    \textbf{Homology ($H_k = Z_k / B_k$)} \\
    Persistent cycles that are not boundaries \\
    ($\text{Betti numbers } b_k = \dim H_k$)
};
\draw[smallarrow] (math_exp.south) -- (homology_def.north); 

\node[mathbox, text width=4.5cm, align=left, right=0.8cm of homology_def] (euler_poincare) {
    \textbf{Euler-Poincaré Theorem:} \\
    $\chi = \sum (-1)^k c_k = \sum (-1)^k b_k$ \\
    \textit{Innately structured by parity}
};
\draw[smallarrow] (homology_def.east) -- (euler_poincare.west);

\node[paneltitle, anchor=west] (panelB_title) at (2, 5.5) {(b) The Parity Hypothesis (Function)};

\node[parity_box_even, anchor=north] (even_parity) at (5, 4.5) { 
    \textbf{Even ($H_0, H_2, \dots$)} \\
    \textbf{Scaffolds} \\
    \textit{\small Static, Integrative Structures} \\
    \textbf{Function: Bind "Wholes"}
};

\node[parity_box_odd, below=0.8cm of even_parity] (odd_parity) {
    \textbf{Odd ($H_1, H_3, \dots$)} \\
    \textbf{Flows} \\
    \textit{\small Dynamic, Recurrent Paths} \\
    \textbf{Function: Run "Processes"}
};

\draw[arrow, dashed] (euler_poincare.east) -- node[above, detail, xshift=0cm, yshift=0.3cm, text width=2cm, align=center] {Leads to functional split} ([xshift=0cm]even_parity.west |- euler_poincare);

\end{tikzpicture}
    \caption{\textbf{The Functional Interpretation of Homological Parity Hypothesis.} (a) The $\partial^2=0$ rule ensures all boundaries (like the edges of a face) are closed, forming a cycle with no endpoints. (b) This consistency leads to a parity-based hypothesis where even-dimensional homology provides stable scaffolds (integrative structures) and odd-dimensional homology provides dynamic flows (recurrent processes) for high-level cognition, including visual perception.}
    \label{fig:fig2}
\end{figure}


We now propose the central thesis: this mathematical parity split is not an accident, but a \textbf{design principle} that the brain exploits for computation. We hypothesize that even- and odd-dimensional homology groups have distinct, complementary functional roles (Fig. \ref{fig:fig2}), as summarized in the following box.

\fbox{\begin{minipage}{0.9\textwidth}
\begin{center}
\textbf{The Homological Parity Principle}    
\end{center}

\noindent The principle posits a fundamental dichotomy in cognitive topology. \textbf{Even-dimensional homology ($\Heven$)} represents the stable, time-invariant \textbf{scaffold, structure, or context} of a system (e.g., a generative model, semantic rules). Conversely, \textbf{Odd-dimensional homology ($\Hodd$)} represents the dynamic, time-varying \textbf{flow, process, or content} that unfolds upon that structure (e.g., an episodic memory, a conscious thought).
\end{minipage}}

\subsection*{Plausible Neural Mechanisms}

The homological framework maps directly onto plausible neurodynamics\cite{gerstner2014neuronal}, resolving the abstraction of topology into specific, known neural processes. The key is to understand that the brain's computational structure is not just spatial, but inherently \emph{spatio-temporal}.
A powerful mechanism is found in \textbf{Polychronous Neural Groups (PNGs)}\cite{izhikevich2006polychronization}, which posits that the brain's fixed axonal transmission delays are not a bug, but a critical computational feature. This single concept powerfully implements both arms of our parity hypothesis:

\begin{itemize}
    \item \textbf{Scaffolds ($\Heven$) $\to$ Spatio-temporal Resonance:} The binding of a percept is the high-energy formation of a stable, analogous to ``standing wave'' (a generalized phase) across a neural population \cite{buzsaki2006rhythms}. This standing wave is not just spatial, but \emph{spatio-temporal}. The brain's matrix of inherent axonal delays forms the temporal scaffold \cite{izhikevich2006polychronization}, a $\Heven$ structure that defines which complex, large-scale resonances are possible. Binding, then, is the \emph{activation} of one of these globally stable, spatio-temporal modes.

    \item \textbf{Flows ($\Hodd$) $\to$ Polychronous Cascades:} The flow is implemented by the PNGs that run on this scaffold. A PNG is a complex $b_1$ cycle (i.e., recurrent trace, $A \to B \to C \to A$) whose very existence is stabilized by the precise delays. A temporal process, like an attentional state or the effortless switch in multistable perception \cite{leopold1999multistable}, is the low-energy phase traversal of this $b_1$ cycle. This is implemented physically as the activation of a single, asynchronous spike cascade. This model is far more powerful than simple sync \cite{strogatz2004sync}, as it allows a single neuron to participate in countless $b_1$ flows, enabling the immense compositional memory capacity the brain requires.
\end{itemize}


\section*{The Computational Objective: From Information to Homology}

Why would the brain's computational architecture adopt such a complex topological structure? We propose that this architecture is not a pre-specified design but the inevitable, emergent result of a single, fundamental information-theoretic objective: the drive to minimize \textbf{context-content uncertainty} \cite{robertson1929uncertainty}.
This fundamental principle posits that a stable, intelligent system must continuously learn and update its internal model to minimize surprise or uncertainty between what it is currently processing ($\Phi$, the \textbf{content}) and the surrounding world-model ($\Psi$, the \textbf{context}). The system's goal is to achieve a state of \emph{informational/topological self-consistency}, where context and content become maximally predictive of one another \cite{keller2018predictive}.

\subsection*{The Duality: Information-Minimization $\Longleftrightarrow$ Topological-Closure}

We propose that the brain's information-theoretic drive and its latent topological structure are two sides of the same coin. A neural world-model achieves a state of \textbf{minimal Content-Context Uncertainty ($\mathcal{U}(\Psi,\Phi) = \min$)} if and only if it is \textbf{topologically self-consistent} ($\chi \approx 0$).
This perfect self-consistency, which is the mathematical fixed point of the CCUP learning rule, requires two conditions:
1) \emph{local closure ($\partial^2=0$):} All local processes and flows ($H_{odd}$) must be self-consistent. There can be no second-order errors or loose ends in the system's dynamic, recurrent operations.
2) \emph{global closure ($H^1(\mathcal{F})=0$):} All local scaffolds ($H_{even}$) must be globally gluable as in sheaf theory \cite{ayzenberg2025sheaf}. The system must be able to seamlessly integrate all its local pieces of context into a single, unified, and coherent world model \cite{ha2018world}, with no gluing obstructions or large-scale contradictions.
In essence, the drive to minimize informational surprise is the very force that builds and selects for a topologically perfect structure. An informationally-optimal brain is one that has eliminated all of its own structural inconsistencies.
In summary, $\chi = \Heven - \Hodd$ is the mathematical justification for our hypothesis. The brain's drive for informational-topological closure \emph{innately splits its computational functions by parity}. It separates the system into two opposing but complementary halves, implying a \textbf{topological parity law}: the brain's computational operators themselves must be segregated by this even/odd divide.

\subsection*{The Functional Split: Parity as Computational Operators}

We propose that the CCUP framework is implemented by a continuous, amortized inference cycle that alternates between these two parity-based operators:
\textbf{1. Even Parity ($H_{even}$) $\leftrightarrow$ Integration \& Context (Scaffold)}
The even-dimensional groups ($b_0, b_2, \dots$) represent \textbf{integrative scaffolds}: \textbf{$b_0$} binds data into connected components (the first level of context); \textbf{$b_2$} represents voids or cavities - the perfect structure for acting as a resonant cavity or \textbf{scaffold}.
\textbf{The Operator ($\mathcal{R}$):} This is the brain's \textbf{contextual retrieval} operator. It is a high-energy binding operation. When new content ($\Phi$) arrives, the $H_{even}$ scaffold catches it, binds it, and integrates it into a stable, global whole (the bound percept, $\Psi$). It's the ``backbone/highway'' that holds and stabilizes the overall pattern. 
\textbf{2. Odd Parity ($H_{odd}$) $\leftrightarrow$ Recurrence \& Content (Flow)}
The odd-dimensional groups ($b_1, b_3, \dots$) represent \textbf{recurrent flows}: \textbf{$b_1$} represents loops or cycles -the mathematical embodiment of a \textbf{recurrent trace}, an oscillation, or a dynamic process.
\textbf{The Operator ($\mathcal{F}$):} This is the brain's \textbf{content bootstrapping} operator. It is a low-energy traversal operation. This operator, $\Phi$, \emph{runs on} the scaffold. It's the branch/flow that propagates information, refines inferences over time, and sustains the latent self (the parity remainder) that moves from one stable state to the next.

We posit that cognition is a dynamical cycle, obeying the Context-Content Uncertainty Principle (CCUP) \cite{robertson1929uncertainty}, that seeks to align these parities through two primary modes like the wake-sleep algorithm \cite{hinton1995wake}. The first is \textit{inference} (waking), a context-before-content cycle where the $\Heven$ scaffold provides top-down predictions to constrain the $\Hodd$ flow. The waking mode directly formalizes hierarchical Bayesian inference \cite{rao1999predictive}, with perception as the convergence of the cycle (Fig. \ref{fig:MAI}), where the $\Hodd$ flow ``snaps'' to the nearest low-energy state on the $\Heven$ scaffold. The second mode is \textit{learning} (sleep) \cite{buzsaki1996hippocampo}, an inverted Structure-before-Specificity (SbS) process. With external sensory noise absent, $\Hodd$ content, dominated by replayed episodic memory traces \cite{wilson1994reactivation}, provides training data to anneal the $\Heven$ scaffold. 
We can abstract the consolidation mechanism \cite{squire2015memory} of the learning mode by the following homological memory model. Let $\gamma_i$ denote the $i$-th memory trace (an $H_{odd}$ flow); it is decomposed as:
$\gamma_i = \sigma + \sum_k^{b_i} a_{ik}\,\beta_k + \partial d_i$,
where $\sigma \in Z_k$ is the context backbone ($\Heven$ scaffold) common to multiple traces; $\beta_k$ are the independent recurrent content loops ($\Hodd$ episodic flow); and $\partial d_i$ is the residual boundary representing transient, unbound noise (e.g., sensory experience). The SbS learning process is a boundary evolution that minimizes this noise ($\partial d_i \to 0$), leaving the stable memory manifold $\sigma + \sum_k a_{ik}\beta_k$ and thereby refining the $\Heven$ scaffold ($\sigma$).

\begin{figure}[h]
    \centering
    \resizebox{0.8\linewidth}{!}{
\begin{tikzpicture}[
    node distance=1.5cm and 2.5cm, 
    module/.style={draw, thick, rounded corners, minimum width=3.8cm, minimum height=1.6cm, align=center},
    op/.style={draw, thick, rounded corners, minimum width=3.2cm, minimum height=1.0cm, align=center, fill=gray!10},
    homology_box/.style={draw, thick, rounded corners, minimum width=3.0cm, minimum height=1.2cm, align=center, fill=blue!5},
    arrow/.style={->, thick, draw=black!70},
    dashedarrow/.style={->, thick, dashed, draw=gray!70},
    label_style/.style={font=\sffamily\small, text=gray!70},
    cycle_label/.style={font=\sffamily\bfseries\small, text=black!80},
    font=\small
]

\node[module, fill=purple!10] (phi_content) at (0, 0) {
    Context \\ \(\Phi_t\) \\
    \textcolor{blue!70}{\sffamily\bfseries\small $H_{odd}$ (Flow)}
};

\node[module, fill=green!10, below=of phi_content] (psi_context) {
    Content \\ \(\Psi_t\) \\
    \textcolor{blue!70}{\sffamily\bfseries\small $H_{even}$ (Scaffold)}
};

\node[op, fill=orange!15, right=of phi_content] (bootstrap_op) {
    \textbf{Bootstrapping Operator} \\
    \textcolor{red!70}{\sffamily\bfseries\small \(\mathcal{F}\)} \\
    Recur / Propagate
};

\node[op, fill=cyan!15, right=of psi_context] (retrieval_op) {
    \textbf{Retrieval Operator} \\ 
    \textcolor{red!70}{\sffamily\bfseries\small \(\mathcal{R}\)} \\
    Integrate / Bind
};

\node[module, fill=purple!5, right=of bootstrap_op] (phi_next) {
    Predictive Update \\ \(\Phi_{t+1}\) \\
    \textcolor{blue!70}{\sffamily\bfseries\small $H_{odd}$ (Flow)}
};

\node[homology_box, fill=gray!10, text width=4cm, right=of retrieval_op] (closure_goal) {
    \textbf{Global Closure} \\ \(\partial^2=0\) (Topological) \\
    $\Leftrightarrow$ Min CCUP (Info)
};

\draw[arrow] (phi_content.east) -- node[above, label_style] {Feeds} (bootstrap_op.west);
\draw[arrow] (psi_context.east) -- node[below, label_style] {Constrains} (retrieval_op.west);

\draw[arrow] (bootstrap_op.east) -- (phi_next.west);
\draw[arrow] (retrieval_op.east) -- (phi_next.west);

\draw[arrow] (phi_next.south) -- (closure_goal.north);

\draw[arrow, dashed] (bootstrap_op.east) -- (closure_goal.west);
\draw[arrow, dashed] (retrieval_op.east) -- (closure_goal.west);

\draw[dashedarrow, bend right=45] (phi_next.north) to node[above, label_style] {Updates Context} (phi_content.north);
\draw[dashedarrow, bend left=45] (phi_next.south) to node[below, label_style] {Updates Content} (psi_context.south);

\node[cycle_label, text width=6cm, align=center] at ($(bootstrap_op)!0.5!(retrieval_op) + (0, 3.cm)$) {
    \textbf{Amortized Inference Cycle} \\
    $\Phi_{t+1}=\mathcal{F}(\Phi_t,\Psi_t),~ \Psi_t=\mathcal{R}(\Phi_{t+1},\Psi_t)$
};

\end{tikzpicture}
}
    \caption{\textbf{Memory-Amortized Inference (MAI) Cycle as a Homological Parity Split}.
The computational model for minimizing content-context uncertainty. The cycle is partitioned by topological parity. Odd-dimensional homology ($H_{odd}$), representing content ($\Phi_t$), is processed by the bootstrapping operator ($\mathcal{F}$). Simultaneously, even-dimensional homology ($H_{even}$), representing context ($\Psi_t$), is processed by the retrieval operator ($\mathcal{R}$). Both operators drive the predictive update ($\Phi_{t+1}$), which in turn updates the system, seeking a state of global closure ($\partial^2=0$).}
\vspace{-0.5cm}
    \label{fig:MAI}
\end{figure}

\section*{Applying the Framework: A Homological Solution to Visual Perception}

The abstract topological parity between scaffold ($\Heven$) and flow ($\Hodd$) finds its most direct biological instantiation in the dual-coding strategy of the mammalian cortex. We propose that the brain distinguishes content from context not just anatomically (i.e., ventral vs. dorsal streams \cite{ungerleider1994and}) but dynamically, through the interplay of rate coding and phase coding \cite{dayan2005theoretical}. This distinction provides a mechanistic explanation for the fundamental constraints of human cognition, including the bistability of perception (fast thinking) and the capacity limits of working memory (slow thinking).

\noindent\textbf{The Parity of Neural Codes}
Standard neurophysiology distinguishes between the firing rate (intensity) and the precise spike timing (phase relative to an oscillation). We map this to the Homological Parity Principle:
1) \textit{``What'' is $\Heven$ (Rate/Amplitude):} The identity of an object must be invariant to time. Therefore, the Ventral stream encodes \textit{content} ($\Phi$) primarily through \textit{rate codes} and synaptic weights. A high firing rate represents the depth or certainty of an $\Heven$ attractor (a static component $\beta_0$), independent of when it occurs.
2) \textit{``Where'' is $\Hodd$ (Phase/Timing):} The location and relation of an object are inherently temporal and dynamic. Therefore, the dorsal stream encodes \textbf{context} ($\Psi$) through \textit{phase codes}. The specific timing of a spike relative to a theta/gamma oscillation \cite{lisman2013theta} defines its position in a sequence or trajectory ($\beta_1$ cycle).
The parity interpretation offers fresh insight into long-standing open problems in visual perception (binding, invariance, and inverse).

\subsection*{The Binding Problem as Scaffold Resonance}

The homological model reframes the binding problem from an unsolvable mystery (``feature-binding is ill-posed'' \cite{di2012feature}) into a specific, answerable, structural question.
The problem feels ill-posed if one looks for a master neuron (i.e., following the single-neuron doctrine \cite{barlow1972single}) or a physical place where binding happens \cite{treisman1996binding}. The homological model offers a new perspective: \emph{don't look for a binder; look for a scaffold}.

\paragraph{Binding is Even-Dimensional Scaffolding}

In our model, the binding of features (color, shape, motion) is the formation of a stable, even-dimensional homological structure (a $b_2$ void).
Here’s how it works:
1)  Distributed Features: The brain's visual pathways process color, shape, and motion information in different, parallel areas, which serve as the inputs.
2)  The Scaffold ($b_2$): The brain's neural architecture contains a vast, pre-existing latent manifold of possible percepts, which is full of $b_2$ (and higher even-dim) holes. Each $b_2$ hole represents a resonant cavity for a specific, coherent whole.
3)  The Binding (Resonance): The simultaneous firing of the color, shape, and motion percept neurons acts like the vibrations on the Chladni plate \cite{zhou2016controlling}. These inputs excite the specific $b_2$ resonant cavity that corresponds to a red-round-moving object.
4)  The Percept (The Generalized Phase): The bound percept (red-round-moving) is the stable, high-dimensional standing wave (the generalized phase from a scalar to a vector) that forms on that $b_2$ scaffold. The features are bound because they are all co-participants in a single, unified resonance. In summary, feature binding is neither ill-posed nor a computation but a structural resonance.

\paragraph{New Perspective Offered by the Parity Solution: Binding for Invariance}

The binding problem actually has two parts: binding features in space and binding them in time. The parity interpretation assigns a different job to each:
1) Even-dim ($b_2$) binds space: The $b_2$ scaffold solves the classic binding problem by integrating the spatial features (e.g., color and shape) into a single object right now. This is the coherent whole.
2) Odd-dim ($b_1$) binds time: The $b_1$ loop solves the temporal binding problem. It is the recurrent trace that sustains the object's identity over time (e.g., ``it is still the same red ball... now it is rolling'').
We need both to solve the binding problem: the $b_2$ scaffold binds the object, and the $b_1$ loop (the latent self or attentional process) latches onto that scaffold and follows it, creating a continuous experience. Our interpretation is consistent with the observation that one can not perceive an object until attending to it (i.e., $b_1$ engagement) \cite{luck1998role}.
Meanwhile, the proposed solution suggests that the $\Heven$ scaffold creates the entire object, which the parity-sorting mechanism must disentangle from its $\Hodd$ flows to achieve invariance, as we will elaborate on next.

\subsection*{Object Invariance as Parity Separation}

How does the visual cortex achieve invariant objective recognition? 
The cortically local subspace untangling hypothesis \cite{dicarlo2012does} describes the step-by-step mechanism (a stack of non-linear transformations) that the brain uses, but it lacks a solid mathematical foundation for developing computational algorithms.
Homological theory can fill this gap by explaining the \emph{structural} goal that the iterated cortical processing mechanism achieves.

\paragraph{Manifold Unfolding is Parity Sorting}


Our reasoning starts with the tangled manifold in early visual areas (like V1 \cite{olshausen2005close}).
That is, a single object's identity (e.g., ``coffee mug'') is hopelessly mixed up with its viewing parameters (e.g., angle, light, size).
From a homological perspective, this tangling is a state where the even-dimensional scaffolds ($\Heven$) (the identity) are completely tangled with the odd-dimensional flows ($\Hodd$) (identity-preserving transformations). Cortically local subspace untangling model\cite{dicarlo2012does} proposes that the ventral stream (V1 $\to$ V2 $\to$ V4 $\to$ IT \cite{miyashita1993inferior}) performs a series of transformations to untangle this manifold.
In homology, this manifold unfolding is the process of homological separation or parity sorting. The brain's transformations are actively factoring out the odd-dimensional flows from the even-dimensional scaffolds.
When reaching the goal state of parity sorting (topological self-consistency $\chi \approx 0$), the $\Heven$ scaffolds (the resonant cavities for the object's identity) are cleanly separated in the latent space from the $\Hodd$ flows (the transformations or paths one can traverse between them). 

\paragraph{Why Switching Can be Effortless in Multistable Perception?}
To better illustrate the role of parity sorting, we use multistable perception  \cite{leopold1999multistable} as an example. The latent space is the manifold $\mathcal{Z}$, the underlying map of all possible perceptual states. The stable percepts (e.g., Vase-vs-Face) are stable attractor states within that space. The homology cycle or scaffold is the pre-defined track that connects these stable states - a 1-dimensional phase circle. For example, phase-0 ($a_k=1$) could be the Vase percept, and phase-$\pi$ ($a_k=-1$) could be the Face percept; but they belong to the same scaffold. Unlike energy-based models \cite{lecun2006tutorial}, which require a high-energy jump due to re-computing the entire percept from scratch, switching between stable states becomes effortless when it is implemented by \emph{phase traversal}: Just taking the pre-built track or flow (the $\beta_1$ loop) and moving along that track from one station to the other (e.g., $0 \to \pi$).

\subsection*{The Inverse Problem as Template Selection}

Unlike Marr's theory \cite{marr2010vision}, homological theory reframes the inverse problem from a combinatorial explosion of calculation (e.g., Minsky's search problem \cite{minsky1967computation}) into a much simpler pattern-matching problem (i.e., low-cost verification via resonance \cite{buzsaki2006rhythms}).
The inverse problem is arguably the most challenging in vision because a 2D retinal image is infinitely ambiguous. Homology solves this by asserting that the brain does not calculate the 3D world from scratch or any intermediate representation, such as Marr's 2.5D sketch.
Instead, the brain has a massive, pre-built library of stable 3D topological templates (its homology groups) from which the 2D input just acts as a key to select the best-fit template.

\paragraph{The Library of Wholes: ($\Heven$ Scaffolds)}

The proposed scaffold-flow model is the central solution to the notoriously challenging inverse problem. The brain's latent manifold ($\mathcal{Z}$) is not empty but rich with stable scaffolds ($b_0, b_2, \dots$) that represent coherent wholes learned from experience. We have the following construction based on the model. 1) Templates: The targeted 3D solutions are $b_2$ scaffolds (or $\Heven$ voids) that are the pre-built concepts (e.g., hollow-sphere, solid-cube, human-face, etc.); 2) Mechanism: The 2D image (the 2-chain input) acts as an excitation pattern or a vibration \cite{buzsaki2004neuronal}, which ``rings'' the latent manifold through the MAI cycle, achieving topological self-consistency.
3) Percept: The brain's 3D percept is the $b_2$ scaffold that resonates most strongly with the 2D input. Just as Helmholtz proposed \cite{von1867handbuch}, unconscious inference, such as visual perception, is just a resonance. The brain guesses some stimulus represents a hollow-ball because the 2D circular input fits that template better than it fits the solid-cube template.
The problem is no longer ill-posed because the brain does calculate from infinite 3D possibilities but picks the best match from a finite, learned library of $\Heven$ scaffolds.

\paragraph{Resolving Ambiguity ($\Hodd$ Flows)}

A homological perspective offers a plausible interpretation of various 3D-related optical illusions, such as the Necker Cube.
The 2D input (the lines of the cube) excites two $b_2$ scaffolds almost equally: 
scaffold A (Cube seen from above) vs.
scaffold B (Cube seen from below).
The brain can't resolve this ambiguity with resonance alone. This is where the $b_1$ flow (recurrent trace) takes over: within the $b_1$ loop: A $b_1$ loop (a dynamic, odd-dimensional flow) connects these two stable $b_2$ scaffolds with different phases.
Following the same parity principle as in multistable perception, the switch in perception is the phase traversal along this $b_1$ loop. The brain's latent self (the $b_1$ flow) latches on to scaffold A for a few seconds, then traverses the loop to latch on to scaffold B.
It is also worth mentioning that the reduction of the inverse problem to a simple verification task is consistent with Mountcastle's universal cortical processing hypothesis \cite{mountcastle1957modality}. As we have recently shown, memory-amortized inference (MAI) via Savitch serialization\cite{savitch1970relationships} is the algorithm that makes this verification incredibly cheap. Even though a brute-force verification would require storing the entire map (the whole state space) in memory, a Savitch-based flow via MAI serialized into hardware. The system only needs to store one tiny piece of data - using train track as an analogy of phase traversal, the phase pointer ($a_k$) simply encodes ``mile marker 27 on Track C.''

\section*{Conclusion}
By deriving the \textbf{Homological Parity Principle} from the fundamental conservation law $\partial^2=0$, we have argued that vision is the dynamic equilibration between two conjugate phases of information: the stable structural scaffold ($\Heven$) and the dynamic contextual flow ($\Hodd$).
This framework provides a unified theoretical language for cognitive science beyond vision.
The homological parity hypothesis in this perspective suggests that the brain did not evolve to solve the inverse, binding, invariance problems in vision separately. It evolved a single, powerful computational engine based on topological self-consistency, and this scaffold-and-flow architecture is the universal solution that applies to all of them and beyond (e.g., motor, auditory, and language).

\bibliography{ref}
\end{document}